\documentclass[conference]{IEEEtran}
\IEEEoverridecommandlockouts
\usepackage{cite}
\usepackage{caption}
\usepackage{amsmath,amssymb,amsfonts}
\usepackage{algorithmic}
\usepackage{graphicx}
\usepackage{textcomp}
\usepackage[dvipsnames]{xcolor}
\def\BibTeX{{\rm B\kern-.05em{\sc i\kern-.025em b}\kern-.08em
    T\kern-.1667em\lower.7ex\hbox{E}\kern-.125emX}}

\RequirePackage{xspace}
\makeatletter
\DeclareRobustCommand\onedot{\futurelet\@let@token\@onedot}
\def\@onedot{\ifx\@let@token.\else.\null\fi\xspace}

\def\eg{\emph{e.g}\onedot}

\def\etc{\emph{etc}\onedot}

\makeatother

\begin{document}

\title{DiffusionAgent\includegraphics*[width=0.038\paperwidth]{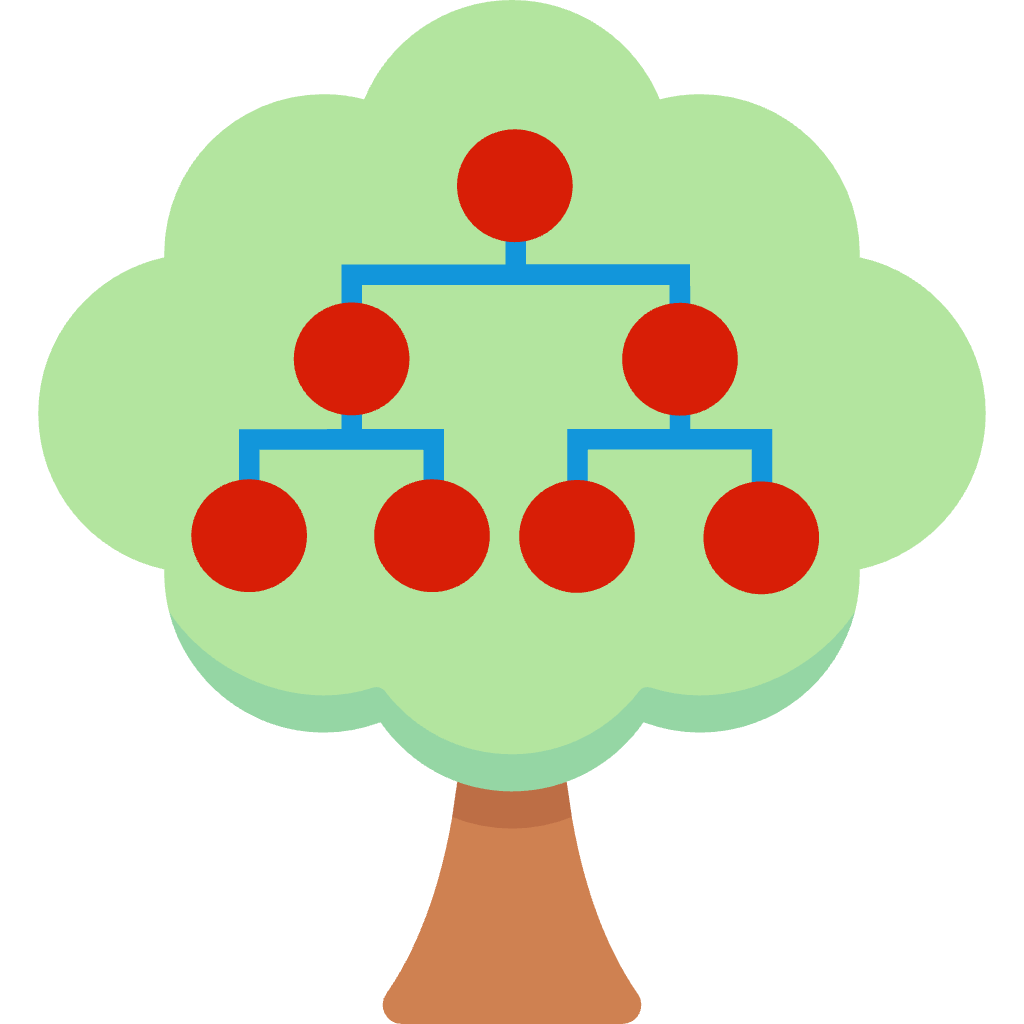}: Navigating Expert Models for Agentic Image Generation}

\author{
Jie Qin$^{1*}$, Jie Wu$^{2*}$, Weifeng Chen$^{2*}$, Yueming Lyu$^{3}$ \\
$^{1}$Meituan $^{2}$ByteDance $^{3}$Nanjing University\\
Project page: \textcolor{blue}{https://DiffusionAgent.github.io
}}

\twocolumn[{%
\renewcommand\twocolumn[1][]{#1}%
\maketitle
\begin{center}
    \centering
    \vspace{-20pt}
    \includegraphics[width=0.9\textwidth]{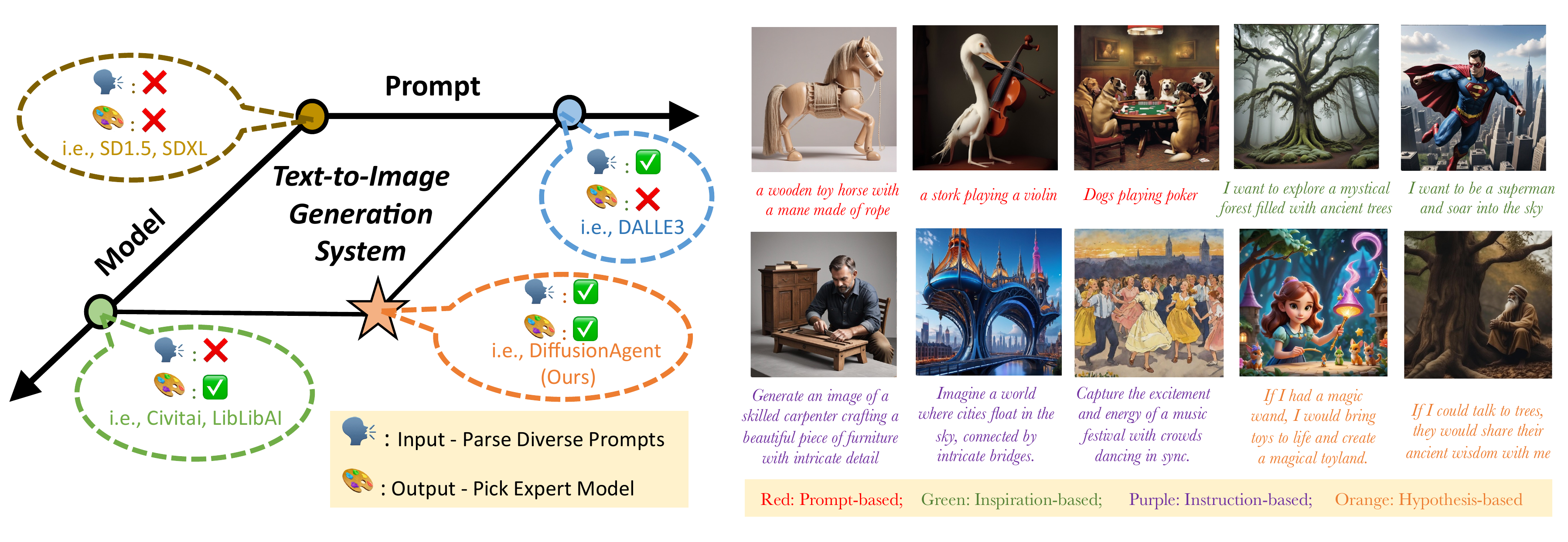}
    \vspace{-8pt}
    \captionof{figure}{We propose a unified generation system DiffusionAgent, which leverages Large Language Models (LLM) to seamlessly accommodate various types of prompts input and integrate domain-expert models for output. Our system is capable of parsing diverse forms of inputs, including \textcolor{red}{Prompt-based}, \textcolor{YellowGreen}{Instruction-based}, \textcolor{violet}{Inspiration-based}, and \textcolor{YellowOrange}{Hypothesis-based} input types. It exhibits the ability to generate outputs of superior quality.}
    \label{fig:head}
\end{center}%
}]

\begin{abstract}

  In the accelerating era of human-instructed visual content creation, diffusion models have demonstrated remarkable generative potential. Yet their deployment is constrained by a dual bottleneck: \textit{semantic ambiguity in diverse prompts and the narrow specialization of individual models}. A single diffusion architecture struggles to maintain optimal performance across heterogeneous prompts, while conventional “parse-then-call” pipelines artificially separate semantic understanding from generative execution. To bridge this gap, we introduce \textbf{DiffusionAgent}, a unified, language-model-driven agent that casts the entire ``prompt comprehension–expert routing–image synthesis" loop into a agentic framework. Our contributions are three-fold: (1) a tree-of-thought-powered expert navigator that performs fine-grained semantic parsing and zero-shot matching to the most suitable diffusion model via an extensible prior-knowledge tree; (2) an advantage database updated with human-in-the-loop feedback, continually aligning model-selection policy with human aesthetic and semantic preferences; and (3) a fully decoupled agent architecture that activates the optimal generative path for open-domain prompts without retraining or fine-tuning any expert. Extensive experiments show that DiffusionAgent retains high generation quality while significantly broadening prompt coverage, establishing a new performance and generality benchmark for multi-domain image synthesis. The code is available at \textcolor{blue}{https://github.com/DiffusionAgent/DiffusionAgent}.

\end{abstract}


\section{Introduction}
\label{sec:intro}

Recent years have witnessed the prevalence of diffusion models\cite{DDPM, SD15, sdxl, dreambooth} in human-interactive image generation tasks, revolutionizing image editing, stylization, and other related tasks.
The first open-source text-to-image diffusion model, known as Stable Diffusion (SD)\cite{SD15}, which has rapidly gained popularity and widespread usage. Various techniques tailored for SD, such as Controlnet\cite{controlnet}, Lora\cite{lora}, further paved the way for the development of SD and foster its integration into various applications.
SDXL\cite{sdxl} is tailored to deliver exceptional photorealistic outputs with intricate details and artistic composition.
Moreover, the impact of SD extends beyond technical aspects. Community platforms such as Civitai, WebUI, and LibLibAI have emerged as vibrant hubs for discussions and collaborations among designers and creators.

Despite making significant strides, current stable diffusion models face two key challenges when applied to realistic scenarios as shown in Fig. \ref{fig:head}:

\begin{itemize}
    \item \textbf{\textit{Model Limitation}}: 
    Stable diffusion models like SD1.5\cite{SD15} are versatile but underperform in specific domains, whereas domain-specific models like SD1.5+Lora excel in targeted sub-fields at the cost of general adaptability.
    
    \item \textbf{\textit{Prompt Constraint}}: 
    During stable diffusion training, text data only includes descriptive captions, yet the models face challenges in achieving optimal performance with diverse prompt, such as instructions and inspirations.

\end{itemize}

\begin{figure*}[t]
    \centering
    \begin{center}
        \includegraphics[width=0.9\textwidth]{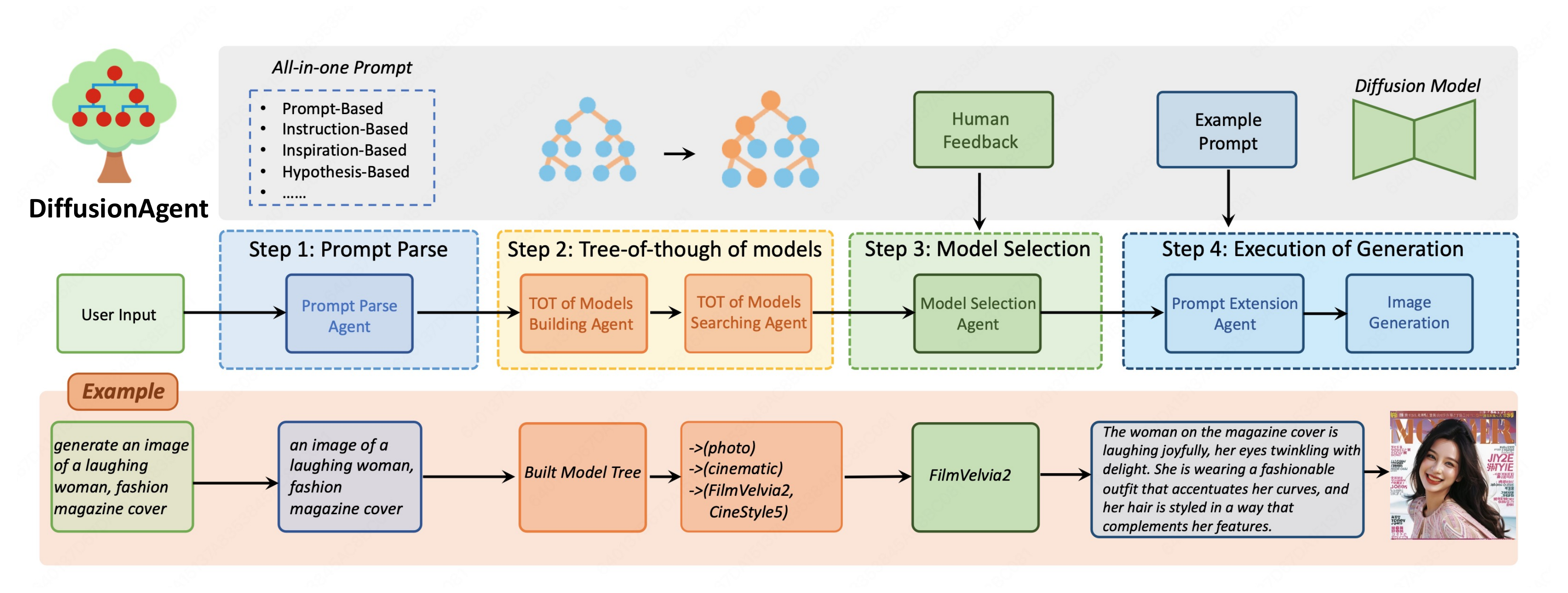}
    \end{center}
        \vspace{-15pt}
    \caption{Overview of \textbf{DiffusionAgent}. The workflow of DiffusionAgent consists of four steps: Prompt Parse, Tree-of-thought of Models of Building and Searching, Model Selection, and Execution Generation. The four steps are shown from left to right and interact with LLM continuously. The upper side shows the detailed process of each step. The lower side shows the example of the whole workflow. }
    \label{fig: framework}
    \vspace{-15pt}
    \end{figure*}

The mismatch between stable diffusion models and real-world applications often results in limited performance, poor generalization, and implementation difficulties. Research has sought to address these issues. While SDXL has improved specific-domain performance, but optimal results are still unattained. Techniques like prompt engineering and fixed templates have been used to enhance input quality and output, yet they fall short of offering a complete solution. This leads us to pose a key question: \textit{Can we create a unified framework to unleash prompt constraint and activate corresponding domain expert model ?}

In order to address the aforementioned question, we propose \textbf{DiffusionAgent}, which  leverages Large Language Model (LLM) ~\cite{instructGPT, Llama, qwen} to offer a one-for-all generation system that seamlessly integrates superior generative models and effectively parses diverse prompts. 
DiffusionAgent constructs a Tree-of-Thought (ToT) structure, which encompasses various generative models based on prior knowledge and human feedback. When presented with an input prompt, the LLM first parses the prompt and then guides the ToT to identify the most suitable model for generating the desired output. 
Furthermore, we introduce Advantage Databases, where the Tree-of-Thought is enriched with valuable human feedback, aligning the LLM's model selection with human preferences.

The contributions of this work can be summarized as:
\begin{itemize}
\item \underline{\textit{New Insight}}: 
DiffusionAgent utilizes a Large Language Model (LLM) as the cognitive engine for its text-to-image generation system, processing various inputs and enabling expert selection for outputs.
\item \underline{\textit{All-in-one System}}: 
DiffusionAgent provides a versatile solution compatible with various diffusion models, unlike existing approaches that are restricted to descriptive prompts, thereby broadening its applicability to different prompt types.
\item \underline{\textit{Efficiency and Pioneering}}: 
DiffusionAgent is notable for its training-free, plug-and-play integration. By incorporating the Tree-of-Thought (ToT) and human feedback, our system enhances accuracy and facilitates a flexible process for aggregating multiple experts.
\item \underline{\textit{High Effectiveness}}:  
DiffusionAgent outperforms traditional stable diffusion models, showcasing significant advancements and offering an all-in-one system that provides a more efficient and effective pathway for community development in image generation.
\end{itemize}

\section{Methodology}

DiffusionAgent is an all-in-one system specifically designed to generate high-quality images for diverse input prompts. Its primary objective is to parse the input prompt and identify the generative model that produces the most optimal results, which is high-generalization, high-utility, and convenient.

DiffusionAgent composes of a large language model (LLM) and various domain-expert generative models from the open-source communities (\eg. Hugging Face, Civitai). The LLM acts as the main controller and maintains the workflow of the system, which consists of four steps: Prompt Parse, Tree-of-thought of Models of Building and Searching, Model Selection with Human Feedback, and Execution of Generation.
The overall pipeline of DiffusionAgent is shown in Fig. \ref{fig: framework}.

\subsection{Prompt Parse}
The Prompt Parse Agent plays a pivotal role in our methodology as it utilizes the large language model (LLM) to analyze and extract the salient textual information from the input prompt. Accurate parsing of the prompt is crucial for effectively generating the desired content, given the inherent complexity of user input. This agent is applicable to various types of prompts, including prompt-based, instruction-based, inspiration-based, hypothesis-based, \etc.

\noindent \textbf{Prompt-based}: The entire input is used as the prompt for generation. For example, if the input is \textit{``a dog"} the prompt used for generation would be \textit{``a dog"}.

\noindent \textbf{Instruction-based}:  The core part of the instruction is extracted as the prompt for generation. For instance, if the input is \textit{``generate an image of a dog"}, the recognized prompt would be \textit{``an image of a dog"}.

\noindent \textbf{Inspiration-based}: The target subject of the desire is extracted and used as the prompt for generation (e.g., Input: \textit{``I want to see a beach"}; Recognized: \textit{``a beach"}).

\noindent \textbf{Hypothesis-based}: It involves extracting the hypothesis condition (\textit{``If xxx, I will xxx"}) and the object of the forthcoming action as the prompt for generation. For instance, if the input is \textit{``If you give me a toy, I will laugh very happily"}, the recognized prompt would be \textit{``a toy and a laugh face"}.

The Prompt Parse Agent helps DiffusionAgent identify core content from prompts, reducing the impact of noisy text. This is essential for selecting suitable generative models and achieving high-quality results.

\subsection{Tree-of-thought of Models}
Following the prompt parsing stage, the subsequent step involves selecting appropriate generative models from an extensive model library to generate the desired images. However, considering the large number of models available, it is impractical to input all models simultaneously into the large language model (LLM) for selection. 
To address this issue and pinpoint the optimal model, we draw inspiration from the Chain-of-Thought (CoT) paradigm~\cite{CoT, fewCoT} and introduce a Tree-of-Thought (ToT) model tree. By leveraging the search capabilities of the model tree, we can narrow down the candidate set of models and enhance the accuracy of the model selection process.

\noindent  \textbf{Constructing the Model Tree using TOT.}
The Tree-of-Thought (TOT) of Model Building Agent is employed to automatically build the Model Tree based on the tag attributes of all models. 
The agent processes the tag attributes of all models to analyze and categorize them into Subject and Style Domains. Style categories become subcategories within the Subject category, forming a two-layer hierarchical tree. Models are assigned to appropriate leaf nodes based on their attributes, completing the model tree. This automatically constructed tree allows for easy extensibility, as new models are seamlessly integrated based on their attributes.

\noindent \textbf{Searching the Model Tree using TOT.}
The Model Tree search process, guided by the Tree-of-Thought (TOT) of Models Searching Agent, aims to find models closely aligned with a given prompt. It uses a breadth-first search, systematically evaluating the best subcategory at each leaf node. Categories are compared against the prompt to find the closest match. This iterative process continues until the final node, yielding a candidate set of models for next selection.

\subsection{Model Selection}
The model selection stage aims to identify the most suitable model for generating the desired image from the candidate set obtained in the previous stage. 
The candidate set, a subset of the model library, includes models closely matching the input prompt. However, limited attribute data from open-source sources complicates precise model selection and detailed information provision to the LLM. To tackle this, we propose a Model Selection Agent that uses human feedback and advanced database technology to align model selection with human preferences.

The advantage database uses a reward model to score model-generated results from 10,000 prompts, storing these scores. For a given input prompt, we calculate its semantic similarity to these prompts and identify the top 5 matches. The Model Selection Agent then retrieves precomputed model performances for these prompts from an offline database, selecting the top 5 models for each. This results in a candidate set of 25 models. The agent intersects the model set with the candidate set from the TOT stage, prioritizing models with higher occurrence probabilities and rankings. These models are chosen for the final model generation.

\begin{figure}[t]
    \centering
    \begin{center}
    \includegraphics[width=0.5\textwidth]{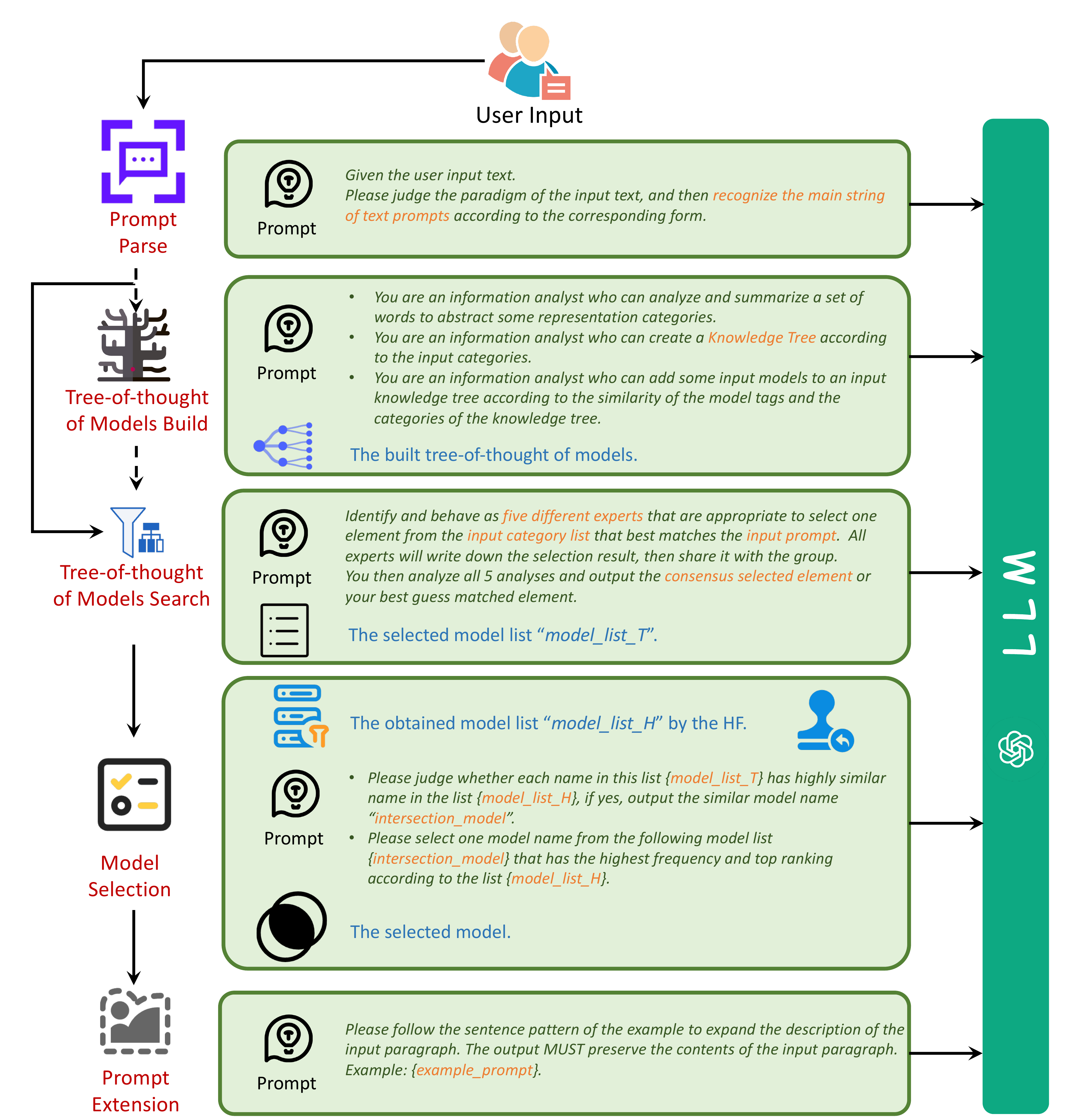}
    \end{center}
    \vspace{-8pt}
    \caption{Details of prompts during interactions with the LLM. Before being inputted into the LLM, the slots ``\{\}''  in figure are uniformly replaced with the corresponding text values.    }
    \label{fig: prompt}
    \vspace{-20pt}
    \end{figure}

\subsection{Execution of Generation}

Once the most suitable model has been selected, the chosen generative model is utilized to generate the desired images using the obtained core prompts.

\noindent \textbf{Prompt Extension.}
To enhance the quality of prompts during the generation process, a Prompt Extension Agent is employed to augment the prompt. This agent leverages prompt examples from the selected model to automatically enrich the input prompt. 
The example prompts and the input prompts are both sent to LLM in the in-context learning paradigm. Specially, this agent incorporates rich descriptions and detailed vocabulary to the input prompts following the sentence pattern of example prompts.
The Prompt Extension Agent enhances it to a more detailed and expressive form, which significantly improves the quality of the outputs.

\begin{figure*}[!t]
    \centering
    \begin{center}
        \includegraphics[width=0.89\textwidth]{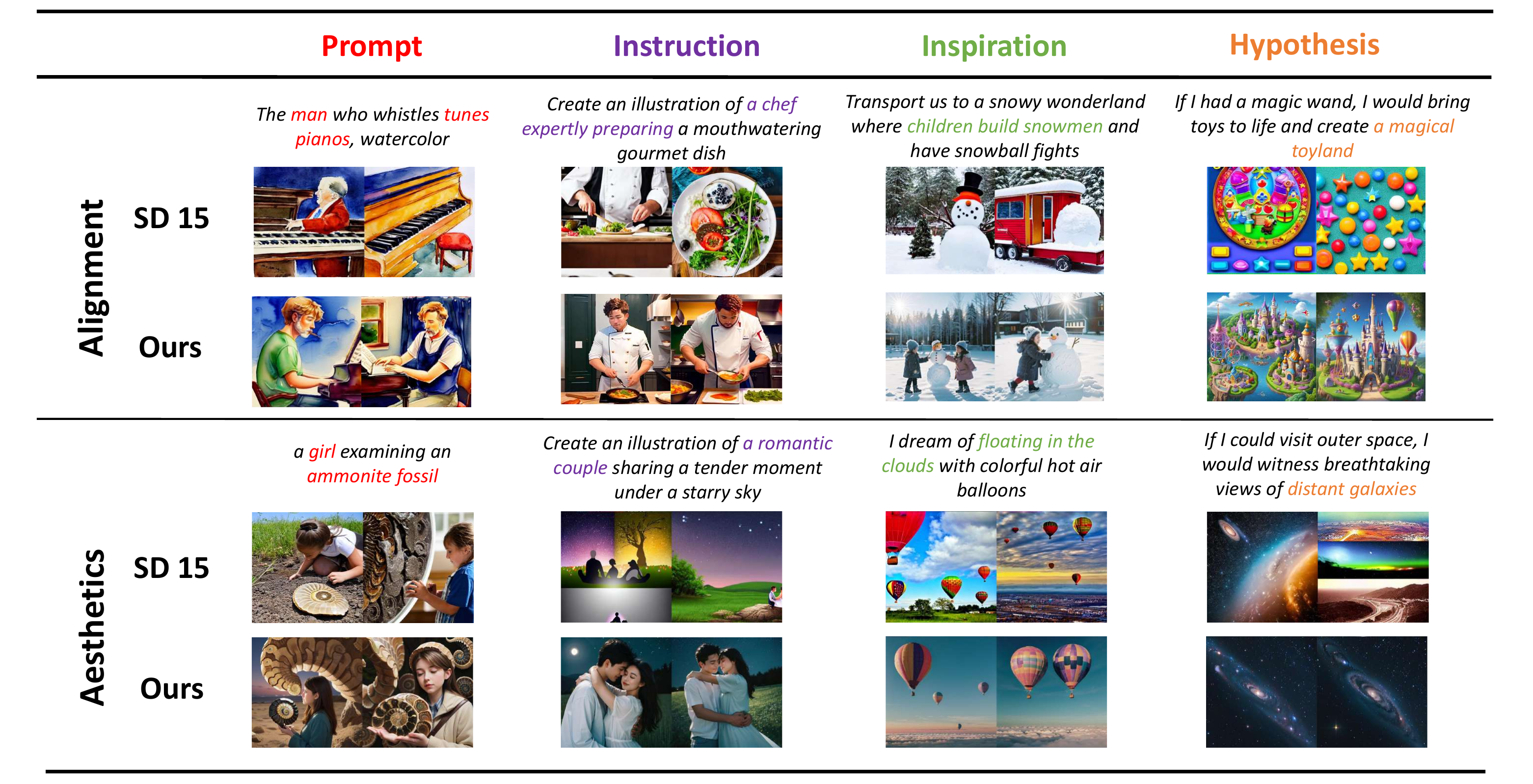}
    \end{center}
        \vspace{-5pt}
    \caption{When comparing SD1.5-based DiffusionAgent with SD1.5\cite{SD15}, it is observed that DiffusionAgent excels in generating more realistic results at a fine-grained level for categories such as humans and scenes. The generated images demonstrate improved visual fidelity, capturing finer details and exhibiting a higher degree of realism compared to SD1.5.}
    \label{fig: compare_base}
    \vspace{-10pt}
    \end{figure*}

\section{Experiments}

\subsection{Settings}

In our experimental setup, the primary large language model (LLM) controller employed was ChatGPT\cite{instructGPT}, specifically utilizing the text-davinci-003 version, which is accessible through the OpenAI API. To facilitate the guidance of LLM responses, we adopted the LangChain framework, which effectively controlled and directed the generated outputs.
For the generation models utilized in our experiments, we selected a diverse range of models sourced from the Civitai and Hugging Face communities. The selection process involved choosing the most popular models across different types or styles available on these platforms. The selected models for different versions of DiffusionAgent are illustrated in Table \ref{tab:models}. The details of prompts during interactions with LLM are shown in Fig. 
 \ref{fig: prompt}.

\begin{table}[t]
    \centering
    \footnotesize
    \caption{The selected public trained models for SD1.5 and SDXL based system. }
      \begin{tabular}{c|c}
        \hline
        \textbf{SD1.5 Based} & \textbf{SDXL Based}\\ 
        \hline
        FilmVelvia2 &  SDXL  \\
        majicmixRealistic & weird-future-fashion \\
        lou & fenrisxl-801Photorealistic \\
        cartoonish & dynavisionXLAllInOneStylized \\
        animemix & sdxlNijiSpecial  \\
        impasto-painting & copaxTimelessxlSDXL1 \\
        CineStyle5 & starcitizen  \\
        dreamlabsoil & lwmirXL \\
        ghibli-style & kandinsky  \\
        dreamshaper & dreamshaperXL \\
        C4D &  \\
        \hline
        \end{tabular}
        \label{tab:models}
    \end{table}

\begin{table}[t]
\centering
\caption{\textbf{Quantitative Results}: We evaluate the aesthetic score using image-reward and aesthetic score compare to ``SD1.5" and ``Random" select expert models to output. }
\label{tab:quantity}
\begin{tabular}{l|ccc}
\hline
Method & Image-reward & Aes score \\ %
\hline
SD1.5 & 0.28 & 5.26  \\
Random & 0.45 & 5.50  \\
DiffusionAgent wo HF & 0.56 & 5.62  \\
DiffusionAgent & \textbf{0.63} & \textbf{5.70} \\
\hline
\end{tabular}
\vspace{-15pt}
\end{table}

\subsection{Qualitative Results}

\subsubsection{Visualization of SD1.5 Version}
To assess the efficacy of our system, we performed a comprehensive evaluation by comparing its generation performance against the baseline method, SD1.5\cite{SD15}. The comparative results are presented in Fig. \ref{fig: compare_base}.
We conducted a detail analysis of four distinct prompt types and compared them along two key dimensions: semantic alignment and image aesthetics.

\begin{figure*}[t]
    \centering
    \begin{center}
        \includegraphics[width=0.89\textwidth]{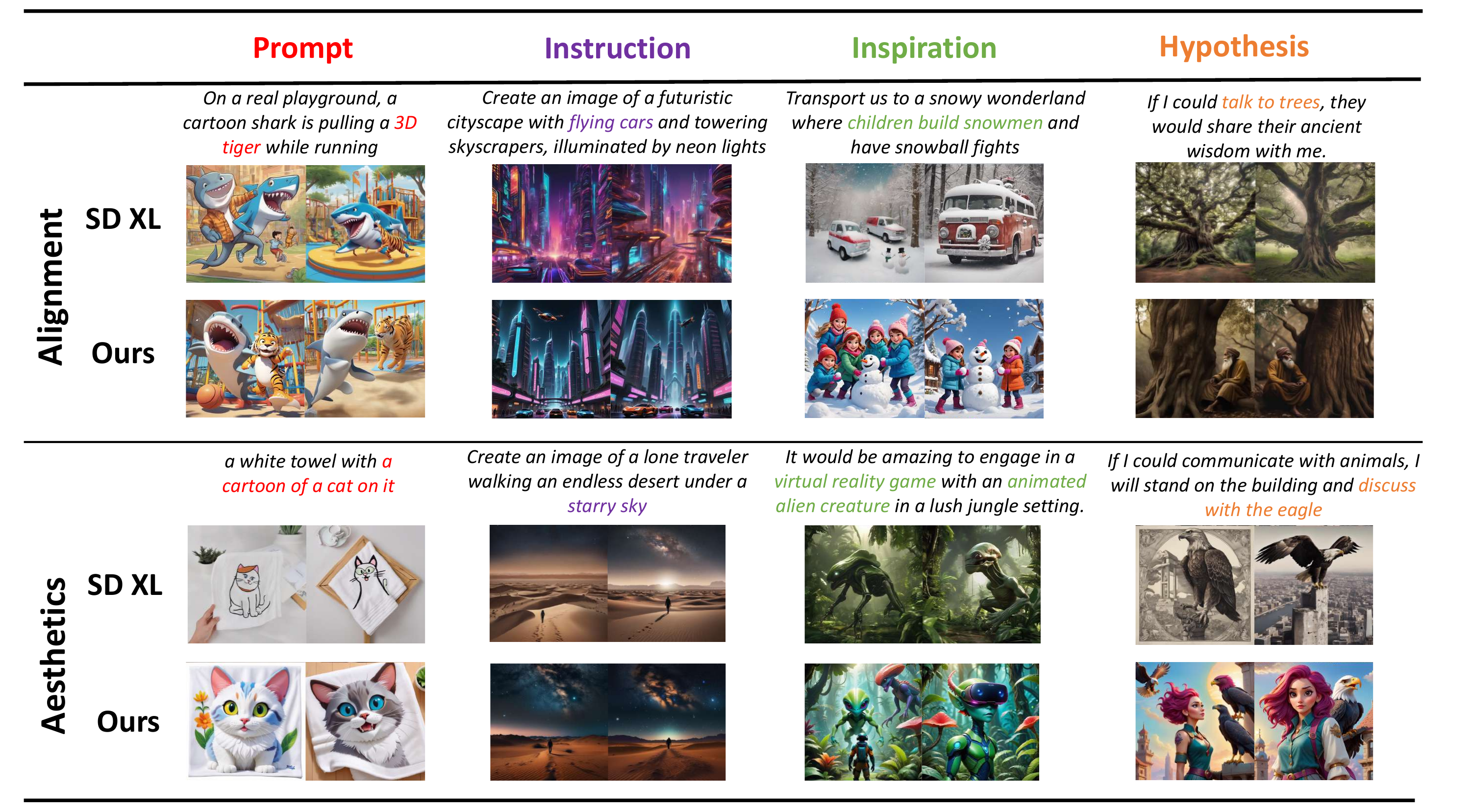}
    \end{center}
        \vspace{-9pt}
    \caption{Comparison of SDXL version of DiffusionAgent with baseline SDXL\cite{sdxl}. All generated iamges are 1024$\times$1024 pixels.}
    \label{fig: compare_xl}
    \vspace{-15pt}
    \end{figure*}

Upon examining the results, two issues with the base model were identified:
i) Semantic Lack: The base model struggles to capture complete semantic information, particularly with prompts involving \textit{``man, chef, children, and toyland"}, focusing too narrowly on specific classes.
ii) Poor Performance on Human-Related Targets: The base model has difficulty generating accurate facial and body details, resulting in lower quality images, especially for \textit{``girl and couple"} prompts.

In contrast, DiffusionAgent effectively addresses these limitations by producing images that capture complete semantic information from input prompts. For example, it successfully represents broader contexts like \textit{``man who whistles tunes pianos"} and \textit{``a snowy wonderland where children build snowmen"}. Additionally, it excels in generating detailed images of human-related objects, as seen with prompts like \textit{``a romantic couple sharing a tender moment under a starry sky"}.

\begin{figure*}[t]
    \centering
    \begin{center}
        \includegraphics[width=0.9\textwidth]{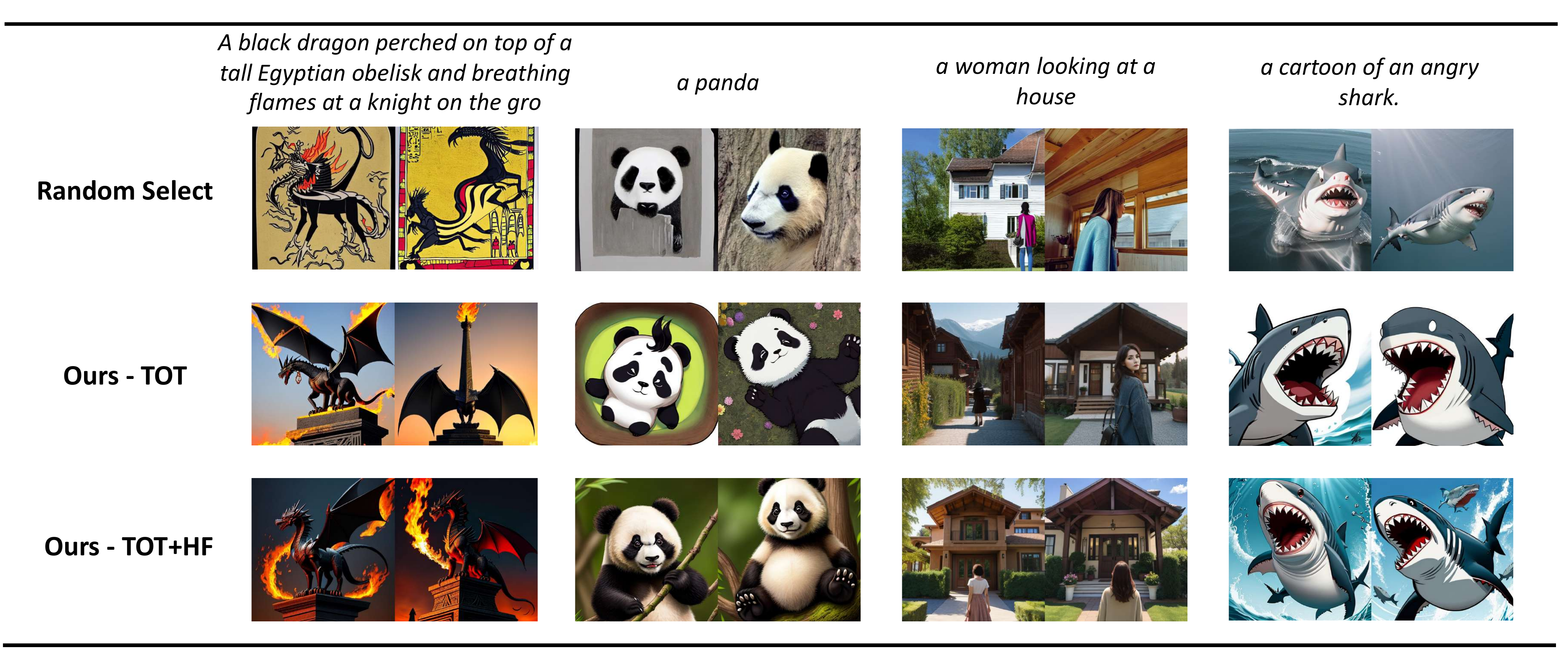}
    \end{center}
    \vspace{-10pt}
    \caption{Ablation study of DiffusionAgent. The random selection is the baseline method for generating images. The TOT or TOT+HF represent the performance of different agents. }
    \label{fig: ablation}
    \vspace{-10pt}
    \end{figure*}

\begin{figure}[t]
    \centering
    \begin{center}
        \includegraphics[width=1.0\linewidth]{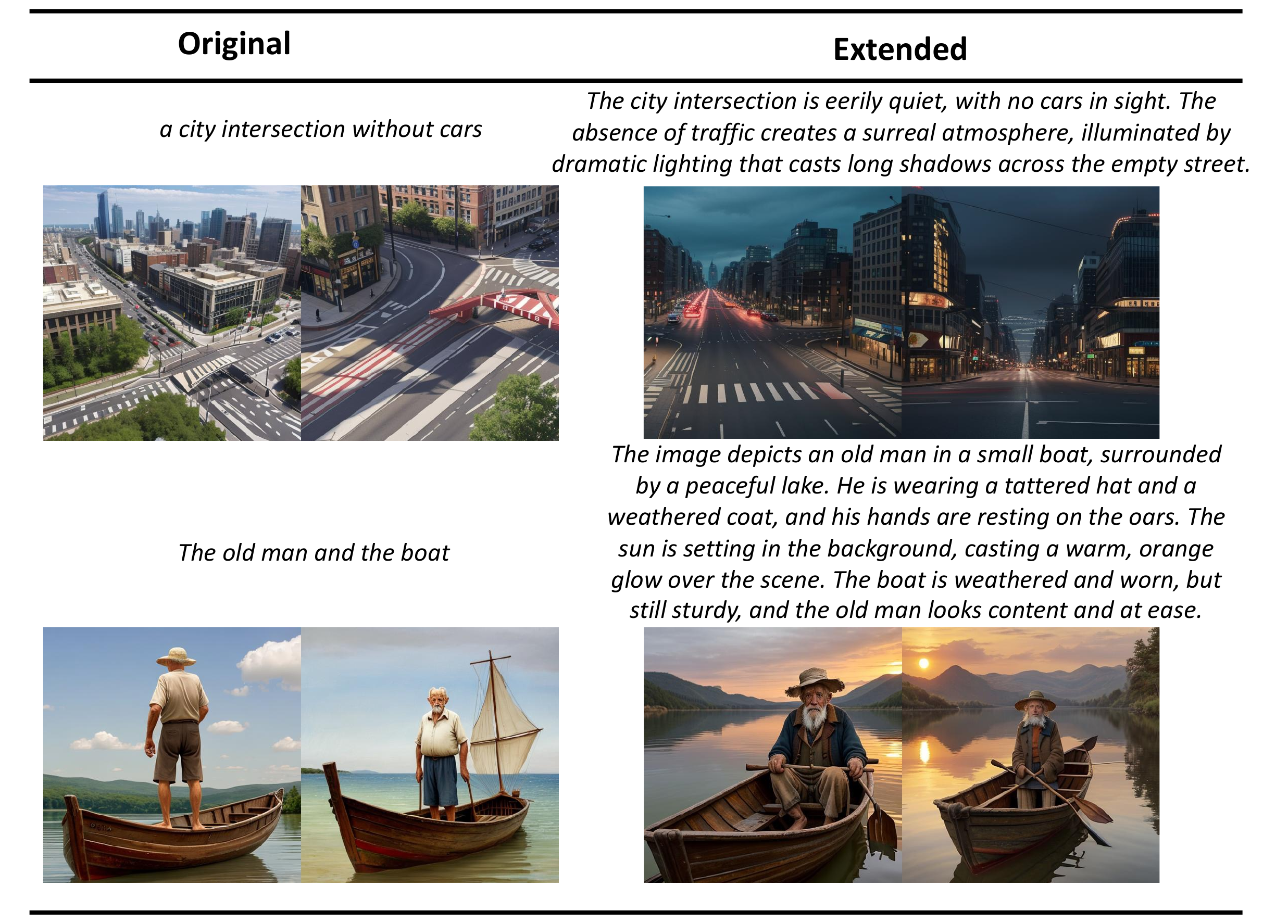}
    \end{center}
    \caption{Ablation of Prompt Extension. It aims to provide the riched prompts that produces higher quality images.}
    \label{fig: boost}
    \vspace{-15pt}
    \end{figure}

\subsubsection{Visualization of SDXL Version.}
With advancements in universal generation models, SDXL\cite{sdxl} has shown promising results. We improved our system by integrating various open-source models based on SDXL. All comparative results are shown in Fig. \ref{fig: compare_xl}, all output images were 1024x1024. The result reveals that SDXL sometimes loses semantic information, as seen in prompts like \textit{``3D tiger"} or \textit{``flying cars"}. In contrast, our system produces more visually appealing images, such as \textit{``a white towel with a cartoon cat"} and a \textit{``starry sky"}.

\subsection{Quantitative Results}

The alignment between user preferences and the quantitative findings presented in Table \ref{tab:quantity} serves as strong evidence of the robustness and effectiveness of DiffusionAgent. To further evaluate the different generated results, we employed the aesthetic predictor and human feedback related reward model. By comparing the effects of our basic version with the baseline model SD1.5, the results in Table \ref{tab:quantity} demonstrate that our overall framework outperforms SD1.5 in terms of image-reward and aesthetic score, achieving improvements of 0.35\% and 0.44\% respectively.

\subsection{Ablation Study}

\subsubsection{Tree-of-Thought and Human Feedback}
To validate the effectiveness of our designed components, we performed a visual analysis of various modules, as shown in Fig. \ref{fig: ablation}. The ``Random" variant represents the random sampling model, which produces many images that lack semantic coherence with the input prompts. In contrast, incorporating the tree-of-thought (TOT) and human feedback (HF) modules significantly enhances image quality, resulting in improved realism, semantic alignment, and aesthetic appeal. This analysis highlights the advantages of our system in selecting superior models through the integration of TOT and HF components.

\subsubsection{Prompt extension}
To evaluate the effectiveness of the prompt extension agent, we compared generation results using original and extended prompts, as shown in Fig. \ref{fig: boost}. The extended prompts provided more comprehensive descriptions, enriching the context for generating visually appealing outputs. Our analysis revealed that extended prompts significantly improved the aesthetics and detail of the generated images, guiding the model to produce outputs with greater fidelity to the desired artistic style. Additionally, the detailed descriptions in the extended prompts enhanced the model's ability to capture intricate nuances, resulting in images that exhibited higher intricacy and refinement compared to those generated with the original prompts.

\subsection{User Study}

\begin{figure}[t]
    \centering
    \begin{center}
        \includegraphics[width=0.48\textwidth]{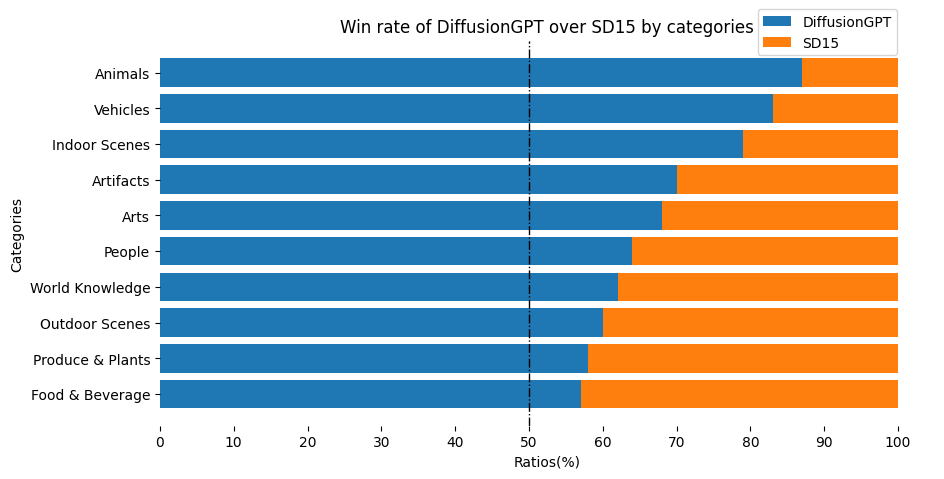}
    \end{center}
        \vspace{-10pt}
    \caption{\textbf{User Study:} Comparing DiffusionAgent with SD1.5. Users  show a strong preference for the expert models selected by DiffusionAgent across all 10 prompt categories.}

    \label{fig: user_study}
    \vspace{-10pt}
    \end{figure}

\begin{figure}[t]
    \centering
    \begin{center}
        \includegraphics[width=0.9\linewidth]{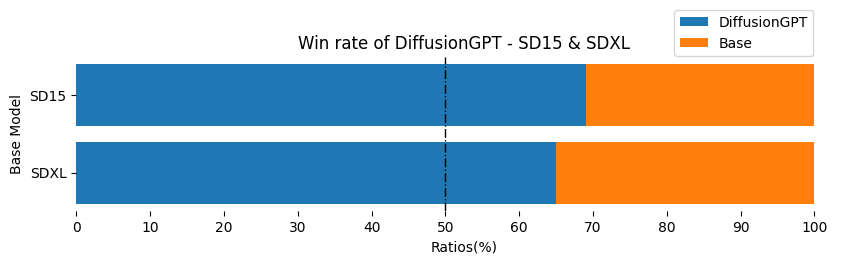}
    \end{center}
        \vspace{-10pt}
    \caption{Comparison of DiffusionAgent-Xl with base model. }
    \label{fig: xl_comp}
    \vspace{-18pt}
    \end{figure}

To assess human preferences for generated images, we conducted a user study comparing our model to a baseline model using 100 randomly selected prompts from PartiPrompts\cite{partiprompt}, generating four images per prompt. Feedback was collected from 20 users who rated the images, resulting in approximately 400 votes for each baseline model (SD1.5 and SD XL). As shown in Fig. \ref{fig: user_study} and \ref{fig: xl_comp}, 
the results consistently favored our model, with users perceiving its outputs as of superior quality versus the baseline.

\section{Related Work}
\label{sec:Related Work}

\subsection{Text-based Image Generation}
Initially, Generative Adversarial Networks (GANs) \cite{GAN, StackGAN} were widely used as the primary approach for Text-based image generation. However, the landscape of image generation has evolved, and diffusion models \cite{DDPM} have emerged as a dominant framework, especially when integrated with text encoders such as CLIP \cite{CLIP} and T5 \cite{T5}, enabling precise text-conditioned image generation \cite{DALLE2,Imgagen,lyu2023deltaedit}. For instance, DAELL-2 \cite{DALLE2} leverages CLIP's image embeddings, derived from CLIP's text embeddings through a prior model, to generate high-quality images. Similarly, Stable Diffusion \cite{SD15} directly generates images from CLIP's text embeddings. Imagen \cite{Imgagen}, on the other hand, utilizes a powerful language model like T5 \cite{T5} to encode text prompts, resulting in accurate image generation. 
To align text-to-image diffusion models with human preferences, recent methods \cite{t2iHF, rldiffusion, image_reward} propose training diffusion models with reward signals. This ensures that the generated images not only meet quality benchmarks but also closely align with human intent and preferences.

\subsection{Large Language Models (LLMs) for Vision-Language Tasks}
The field of natural language processing (NLP) has witnessed a significant transformation with the emergence of large language models (LLMs) \cite{instructGPT, PaLM, Llama}, which have demonstrated remarkable proficiency in human interaction through conversational interfaces. To further enhance the capabilities of LLMs, the Chain-of-Thought (CoT) framework \cite{CoT, fewCoT, zeroCoT, least2more} has been introduced. This framework guides LLMs to generate answers step-by-step, aiming for superior final answers.
Recent research has explored innovative approaches by integrating external tools or models with LLMs \cite{ Toolformer, viperGPT, visChatGPT, huggingGPT}. For example, Toolformer \cite{Toolformer} empowers LLMs with the ability to access external tools through API tags. Visual ChatGPT \cite{visChatGPT} and HuggingGPT \cite{huggingGPT} extend the capabilities of LLMs by enabling them to leverage other models to handle complex tasks that go beyond language boundaries. 
Drawing inspiration from these endeavors, we embrace the concept of LLMs as versatile tools and leverage this paradigm to guide  T2I models to generate high-quality images.

\section{Conclusion}
We propose DiffusionAgent, a one-for-all framework that seamlessly integrates superior generative models and efficiently parses diverse prompts. By leveraging Large Language Models (LLMs), DiffusionAgent gains insights into the intent of input prompts and selects the most suitable model from a Tree-of-Thought (ToT) structure. This framework offers versatility and exceptional performance across different prompts and domains.
To sum up, DiffusionAgent is training-free and can be easily integrated as a plug-and-play solution, offers an efficient and effective pathway for community development.

\bibliographystyle{IEEEbib}
\bibliography{ref}

\end{document}